\documentclass[letterpaper, 10 pt, conference]{ieeeconf}  

\IEEEoverridecommandlockouts                              

\usepackage{graphics} 
\usepackage{epsfig} 
\usepackage{mathptmx} 
\usepackage{times} 
\usepackage{amsmath} 
\usepackage{amssymb}  
\usepackage[T1]{fontenc}
\usepackage{arydshln}
\usepackage{hyphenat}

\title{\LARGE \bf
Diminishing Domain Bias by Leveraging Domain Labels in Object Detection on UAVs
}

\author{Benjamin Kiefer$^{\dagger}$, Martin Messmer$^{\dagger}$ and Andreas Zell \\
University of Tübingen
\thanks{$\dagger$ These two authors contributed equally.}
\thanks{Email:  {\tt\small forename.surname@uni-tuebingen.de}}%
\thanks{This work has been supported by the German Ministry for Economic
Affairs and Energy, Project Avalon, FKZ: 03SX481B.}%

}

\begin{document}

\maketitle
\thispagestyle{empty}
\pagestyle{empty}


\begin{abstract}
Object detection from Unmanned Aerial Vehicles (UAVs) is of great importance in many aerial vision-based applications. 
Despite the great success of generic object detection methods, a significant performance drop is observed when applied to images captured by UAVs. This is due to large variations in imaging conditions, such as varying altitudes, dynamically changing viewing angles, and different capture times. These variations lead to domain imbalances and, thus, trained models suffering from domain bias. We demonstrate that domain knowledge is a valuable source of information and thus propose domain-aware object detectors by using freely accessible sensor data. By splitting the model into cross-domain and domain-specific parts, substantial performance improvements are achieved on multiple data sets across various models and metrics without changing the architecture. In particular, we achieve a new state-of-the-art performance on UAVDT for embedded real-time detectors. Furthermore, we create a new airborne image data set by annotating 13,713 objects in 2,900 images featuring precise altitude and viewing angle annotations.
\end{abstract}


\section{Introduction}

Object detection from Unmanned Aerial Vehicles (UAVs) has developed to an important line of research due to its impact on many application areas, such as traffic surveillance, smart
cities and search and rescue \cite{zhu2018visdrone_test,menouar2017uav,lygouras2019unsupervised,mishra2020drone}. 

While generic object detection
has improved drastically lately \cite{zhao2019object}, object detection on images captured from UAVs still poses great challenges \cite{zhu2020vision}.
Among these, the variation across domains is particularly challenging.

\begin{figure}
	
	\centering
	\centerline{\includegraphics[width=8.5cm]{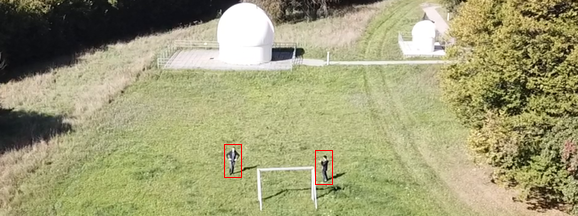}}
		
	\centerline{\includegraphics[width=8.5cm]{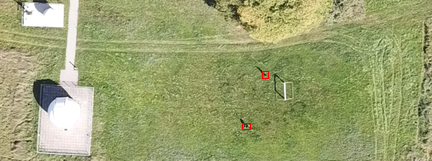}}

	\caption{Example images of the dataset POG, showing the same scenery taken from different perspectives (top: 10m, 10$^{\circ}$, bottom: 100m, 90$^{\circ}$).}
	\label{fig:ppl}
    \vspace{-0.5cm}
\end{figure}

For example, an object detector encounters images taken from varying altitudes. Therefore, the scales of objects vary enormously, often ranging from 10 pixels to over 300 \cite{varga2021seadronessee}. In lower altitudes, objects are captured with more detail while in higher altitudes, more objects appear, but blurrier. 
Furthermore, modern UAVs are equipped with so-called gimbal cameras \cite{rajesh2015camera}. These allow for capturing objects with various viewing angles (pitch axis). Moreover, a UAV's roll axis introduces yet more variation.  As a result, objects are captured with diverse aspect ratios and orientations. In particular, top-down views often result in ambiguous object appearances, such as distinguishing between a car or a van. 

Many more factors influence objects' appearances. These include but are not limited to: variations in weather and time, both affecting the illumination of objects; GPS location; camera sensor. Examples for the individual factors might be: images during rain vs. at sunny weathers; at day vs. at night; different backgrounds resulting from images taken in cities vs. rural areas; lens distortions from different cameras.

These variations become more critical when the interplay with different domains is considered. For example, in Fig. \ref{fig:ppl} the very same scenery is shown from altitudes 10m and 100m, respectively, and from viewing angles 10$^{\circ}$ (nearly horizontally facing) and 90$^{\circ}$ (top-down), respectively. 

In contrast, many traditional data sets in other applications consist of less restricted-view data, such as COCO \cite{lin2014microsoft} for everyday objects, KITTI \cite{geiger2013vision} for autonomous driving and DOTA \cite{xia2018dota} for remote sensing. Therefore, models trained on these data sets do not have to take the aforementioned domain variations into account.

Ultimately, the goal of object detection from UAVs is to detect objects in all of the considered domains. However, data sets are commonly unbalanced with respect to different domains (see Figure \ref{table:visdrone-count}). Therefore, trained models are biased towards over-represented domains while failing to perform well in under-represented domains. As a result, even state-of-the-art models are underoptimized in the latter domains, as will become clear in Section \ref{sec:experiments}. 

In part, this is a consequence of using the commonly used metric average precision.
This domain-agnostic metric favors models, which perform well in over-represented domains but may fail in others. 

Motivated by these observations, we propose to leverage domain labels to alleviate this bias.  While domain information is implicitly encoded in the captured images, it is also explicitly available from the UAVs' sensors: the altitude of the aircraft can be retrieved from the onboard GPS or barometer, the viewing angle from the gimbal pitch angle of the camera, and the time from an onboard clock. We propose to use these domain labels to train so-called expert models. These experts adapt to their respective domains to capture the domain-specific features. This multi-domain learning approach is in contrast to domain adaptation, which aims to eliminate these recognized types of domain. It is furthermore different from multi-task learning as we try to solve the same task across all domains. We show that these experts prove highly effective and efficient across various models, data sets and metrics.

In summary, our contributions are threefold:\vspace*{-2mm}\begin{itemize}
    \item We analyze domain imbalance in three UAV object detection data sets and their effects on the overall model performance.  We also propose a simple domain-sensitive metric to capture domain specific particularities.
    \item We propose a simple method, which leverages domain knowledge, to alleviate domain bias. We show that using this method we can significantly outperform domain-agnostic models without sacrificing speed. Further, we analyze the method on two established UAV object detection data sets. 
    \item  We capture and annotate a UAV object detection data set dubbed PeopleOnGrass (POG). We show that more precise domain labels can improve detection accuracy even further.
\end{itemize}

\section{Related Work}
\label{sec:relatedwork}


Deep learning-based object detection can roughly be divided into two categories: accurate two-stage detectors, like Fast R-CNN or Faster R-CNN \cite{ren2016faster}, and the much faster, but less accurate one-stage detectors such as YOLO \cite{redmon2018yolov3} or
EfficientDet \cite{tan2020efficientdet}. While there is a large amount
of research towards improving these object detectors, much
of the research community focuses mainly on popular benchmarks, such as MS COCO \cite{lin2014microsoft}.

While research fields such as remote sensing used geo-spatial image data sets (e.g. satellite data), they are not that useful for object detection from UAVs since they employ very
low pixel per centimeter resolutions and vary very little in
their altitude and angle information \cite{li2018hsf}. Furthermore, a common practice in object detection from UAVs is still to use off-the-shelf detectors \cite{zhu2018visdrone}.

With the release of the UAVDT \cite{du2018unmanned} and VisDrone \cite{zhu2018visdrone} data sets, several works develop models specifically aimed at object detection from UAVs \cite{vsevo2016convolutional,sommer2017fast,ding2018learning}. Many works focus on detecting small or clustered objects \cite{hong2019patch,yang2019clustered}.

With \cite{bashmal2018siamese}, the concept of domains enters the field of object detection from UAVs, where a Siamese-GAN is introduced to learn invariant feature representations for labeled and unlabeled aerial images from two different domains. However, such a
domain adaptation method's focus is to adapt the model from a fixed
source domain to a fixed target domain. Fine-grained domains are utilized by \cite{wu2019delving}, where adversarial losses are employed to disentangle domain-specific nuisances. However, the training is slow and unstable, and domain labels are ignored at test time.
Expert models are proposed in \cite{lee2019me} to account for objects with particular shapes (horizontally/vertically elongated, square-like). Since no domain labels are used in this work, they are formulated as a model ensemble too expensive to employ in multiple domains. A multi-domain learning approach for object detection is investigated in \cite{wang2019towards}, where the focus is on learning from multiple distinct data sets. 
Transfer learning \cite{zhuang2020comprehensive} is different in that it generally aims to learn invariant representations, whereas multi-domain learning preserves the domain-specific representations.

As opposed to the aforementioned works, we aim to leverage freely available environmental data from the drones' sensors. We try to leverage these so far overlooked domain labels at training and runtime to reduce the domain bias induced by highly imbalanced data sets.

\section{Analyzing Domain Imbalances}

In the following two subsections, we analyze domain imbalances and their consequences in two of the most popular UAV object detection data sets. First, we consider imbalances in the training set and then in the testing set.

\subsection{Domain Imbalances in the Training Set}

\begin{center}

    \begin{table}
    \vspace{1.5mm}
\caption{Available domain labels in the data sets VisDrone and UAVDT and its ranges. Note that the ranges have been estimated by visual inspection since they have not been reported. }
    \label{table:which-meta}

	\centering
		
			\begin{tabular}{c|c|c}
			Domain type & Domain name & Estimated ranges\\
			\hline
			 Altitude & \begin{tabular}{c}
			      high (H)\\
			      medium (M)\\
			      low (L)
			  \end{tabular} &
			 \begin{tabular}{c}
			      80-100m  \\
			      30-80m  \\
			      0-30m 
			  \end{tabular} \\
			  \hline
			  Angle & \begin{tabular}{c}
			      bird-view (B) \\
			      acute angle (A) \\
			      
			  \end{tabular} &
			 \begin{tabular}{c}
			      70-90$^\circ$   \\
			      0-70$^\circ$  \\

			  \end{tabular} \\
			  \hline
			 			  Time & \begin{tabular}{c}
			      day (D)\\
			      night (N)\\
			      
			  \end{tabular} &
			 \begin{tabular}{c}
			      6am-10pm  \\
			     10pm-6am   \\

			  \end{tabular} \\

		\end{tabular}

\end{table}

\end{center}

Imbalance problems in data-driven object detection have been known for a long time. However, most of the literature focuses on class, scale, spatial and objective imbalances \cite{oksuz2020imbalance}. In contrast to many other applications areas, data in object detection from UAVs is versatile with respect to environmental domains. 

So far, we loosely used the term 'domain' to depict a particular environmental state a UAV is in at the time of image capture. Some of these states give rise to some of the imbalances mentioned above: Altitude imbalances give rise to scale imbalances as object sizes directly correlate with the altitude an image is captured at. Also, foreground-background imbalances are affected by the altitude. Viewing angle imbalances give rise to spatial and aspect ratio imbalances. However, there might be many other domain imbalances that may not directly relate to the aforementioned imbalances, such as lighting imbalances caused by the time or weather. 

However, it is not clear what separates one domain from another. In fact, many environmental factors are continuous, such as the altitude or angle an image is captured at. Nevertheless, in current UAV object detection data sets, only coarse domain labels are reported. Two of the most established data sets, UAVDT and VisDrone, feature domain labels with coarse information about altitude, viewing angle and time as depicted in Table \ref{table:which-meta}. Although these divisions seem arbitrary, they already help distinguish features in different domains, as will be seen in Section \ref{sec:experiments}.  

\begin{center}

\begin{figure*}

	
	\includegraphics[trim={0 0 1cm 0},clip,width=8.8cm]{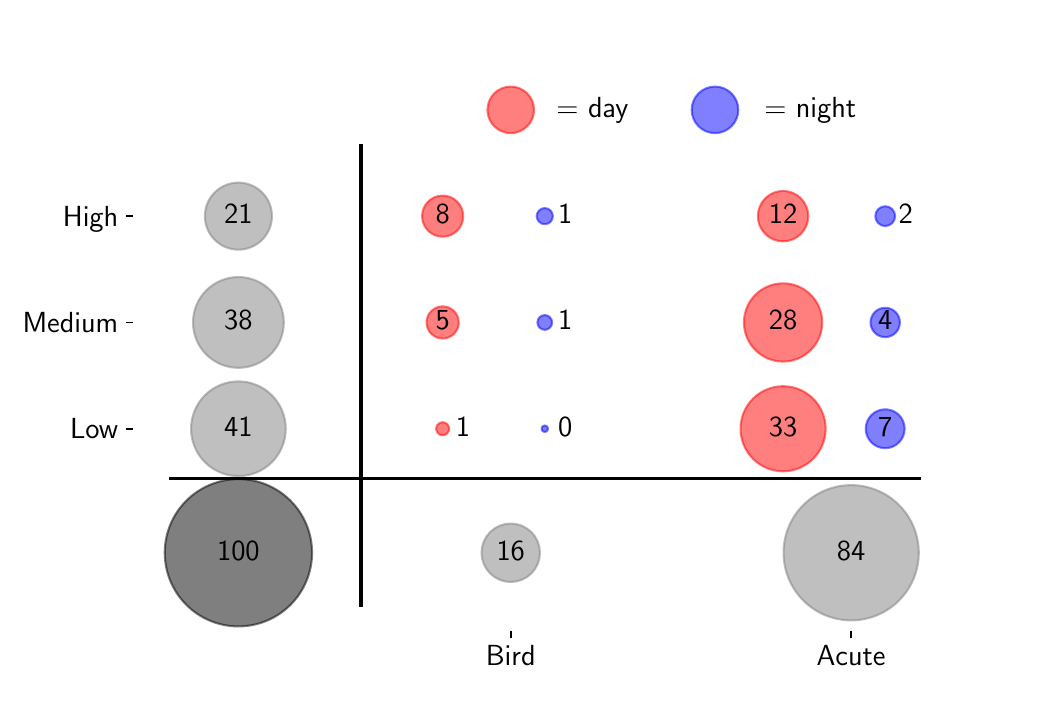}
	\includegraphics[trim={0 0 1cm 0},clip,width=8.8cm]{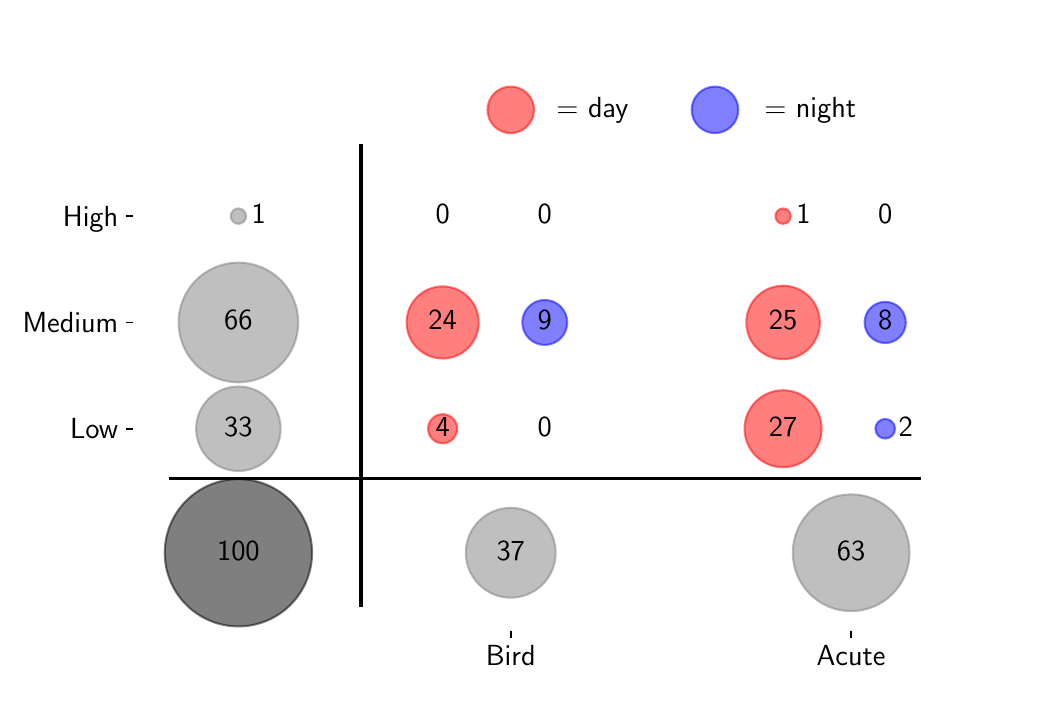}

	\caption{Distribution of objects across domains in the VisDrone (left) and UAVDT (right) training set. The lower left circle represents the size of the whole data set (100\%), the other circles the relative size to it (rounded to the closest integer).
		The domains are high, medium, low, bird view, acute viewing angle, day and night and combination thereof. Best viewed in color.}
	\label{table:visdrone-count}

\end{figure*}

\end{center}

The large amount of varying domains causes data sets to be highly unbalanced with respect to these domains. Figure \ref{table:visdrone-count} shows the number of labeled objects in every domain for the UAVDT and VisDrone training sets. Note that a domain is a combination of one or more influencing variables. For example, the domain+ 'high' (H) + 'bird view' (B) + 'night' (N) in VisDrone contains 4,120 objects. Furthermore note, that we deliberately compared the number of objects and not the number of images because common object detection losses are back-propagated for every object instance -- as opposed to every image. 

In both data sets, many domain imbalances exist. For example, in both data sets, there are fewer labeled objects at night than at day. Both data sets show most objects from a horizontal viewing angle as opposed to from bird-view. These imbalances can be quite large. For example, in VisDrone, the domain H+B+N contains roughly only 1\% ($\approx 4,120/343,205$) of objects, whereas the domain L+A+D contains roughly 33\% ($\approx 114,504/343,205$). Even more extremely, in UAVDT, there are no objects in H+B. 

These domain imbalances result in models being biased towards the over-represented domains. In turn, this may hamper models to predict objects in every domain accurately. In Section \ref{sec:method}, we aim to propose a simple model family to diminish these biases.

\subsection{Domain Imbalances in the Testing Set}
\label{sec:metric}

While domain imbalances in the training set cause a trained model to be biased towards the over-represented domains, domain imbalances in the testing set cause a trained model to be rewarded for that behavior. If we want to faithfully measure the performance of an object detector across domains equally, we ought to include this in the corresponding metric. However, the conventional metric 'mean average precision' (mAP) does not capture the concept of a domain. Indeed, it is designed to be a general-purpose metric that weighs precision and recall. It is the area under the precision-recall curve averaged over all classes $c\in \{1,\dots,C\}$ defined as follows:
\begin{equation}
    \text{mAP} := \frac{1}{C}\sum_{c=1}^C \text{AP(c)}:=\frac{1}{C}\sum_{c=1}^C \int_0^1 p_c(r) \text{d}r,
\end{equation}
where $p_c(r)$ denotes the precision for class $c$ for a recall value $r$. True positives are determined by measuring the intersection-over-union (IoU) of the predicted bounding box and the ground truth. The threshold varies across data sets. Without any subscript, the value denotes the average value over thresholds from 0.5 to 0.95 in steps of 0.05 \cite{lin2014microsoft}. Because there are only finitely many predictions, the integral simplifies to a sum over the ordered object predictions.

\begin{figure}

	\centering
	
    \centerline{\includegraphics[width=8.5cm]{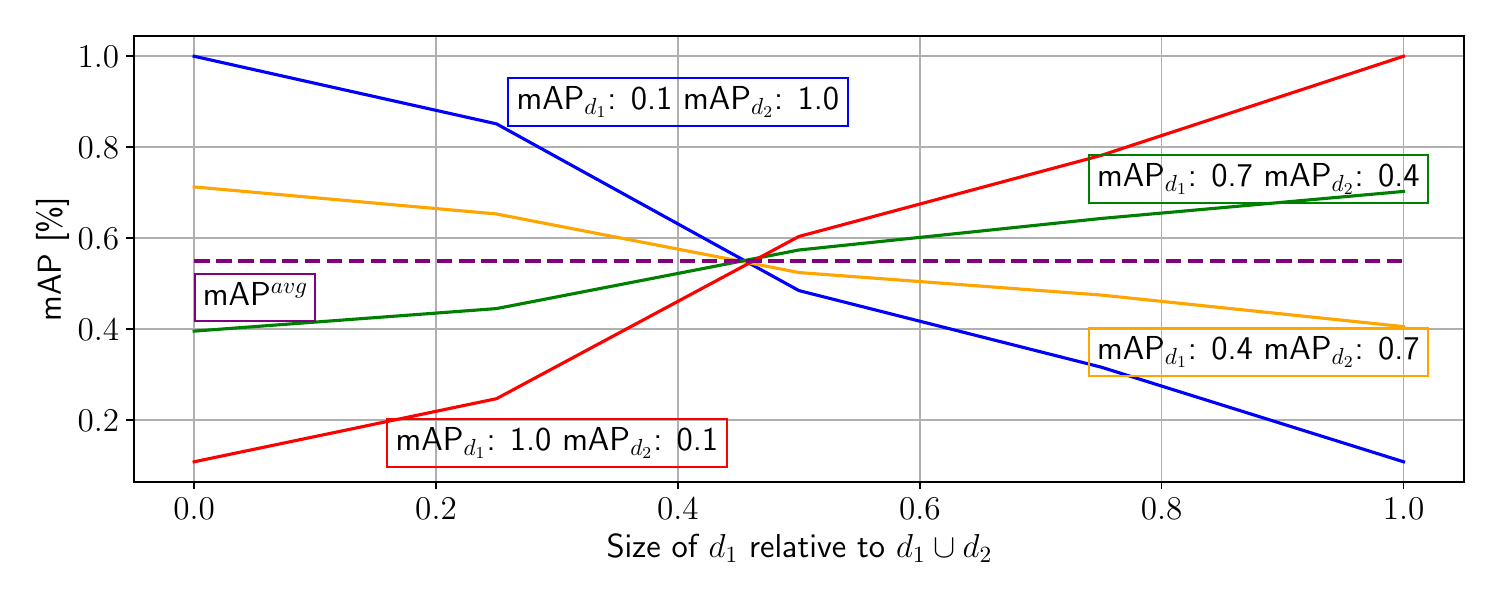}
	}
	
	\caption{Hypothetical mAP values for a two-domain UAV object detection data set. mAP$^\text{avg}$ is the average mAP over both domains as defined in equation \ref{equ:mapavg}. Best viewed in color.}
	\label{fig:mAP_plot}

\end{figure}

To illustrate the severeness of mAP being domain agnostic, consider the following toy example: Suppose we have two distinct domains $d_1$ and $d_2$ in our UAV object detection data set. Let mAP$_{d_1}$ and mAP$_{d_2}$ be the mAP scores of a model trained on all data but evaluated only on $d_1$ and $d_2$, respectively. Denote by $s\in [0,1]$ the size of $d_1$ relative to the size of the whole data set $d_1\cup d_2$. In Figure \ref{fig:mAP_plot}, we plot the mAP on $d_1\cup d_2$ as a function of $s$ for certain fixed values of mAP$_{d_1}$ and mAP$_{d_2}$. Note that these curves depend on the distribution of true/false positives, true/false negatives and scores of the predictions and are therefore not unique.

From that hypothetical example, it is evident that small domains contribute very little to the overall mAP score. For example, consider the blue curve. In this case, mAP$_{d_1}=0.1$ and mAP$_{d_2}=1$. If the size of $d_1$ is less than 1/4 of the whole data set size, the overall mAP still is above 80\%. This leads to overestimating models that just perform well on over-represented domains and underestimating models that perform well on under-represented domains.

Ideally, a UAV object detection data set is roughly balanced with respect to domains. However, as we saw in the subsection before, this condition often is violated. Therefore, the only way to obtain models that are robust across domains is to incorporate this domain performance into the metric. We propose to use the simple domain-averaged metric
\begin{equation}
\label{equ:mapavg}
    \text{mAP}^\text{avg} := \frac{1}{D} \sum_{j=1}^D\text{mAP}_{d_j},
\end{equation}
where mAP$_d$ denotes the mAP on domain $d\in \{d_1,\dots,d_D\}$. To obtain well-calibrated models, we evaluate on both, mAP and mAP$^\text{avg}$. Note that it is a user question of how to weigh each domain. Non-uniform weightings of domains are possible as well. However, we argue that a priori all domains should be weighted equally to allow for cross-domain robust models. In the example from before, the dashed purple curve depicts mAP$^{\text{avg}}$, which is independent of the the sizes of each domain.

\section{Multi-Domain Learning Approach}
\label{sec:method}

For a fixed model architecture, learning from multiple domains is inherently a trade-off. Large domains cause the model to be biased towards these domains. Our goal is to diminish this bias by leveraging freely available domain labels in a multi-domain learning setting. 

In multi-domain learning, image samples $\{x_j\}$ with corresponding bounding box annotations $\{y_j\}$ are accompanied by a discrete domain indicator $d_x\in \{d_1,\dots,d_D\}$ (which also is available at test time), such that a training sample is $(x_j,d_{x_j} ,y_j)$ and a test sample is $(x_j,d_{x_j})$. In particular, that means, we can leverage this domain information $d_x$ at test time, which is the key to our expert models.

Motivated by \cite{caruana1997multitask}, we propose a multi-head architecture. Given a general object detector model, we share earlier layers across all domains and leave later layers domain-specific. This approach follows the empirical observations that earlier layers extract lower-level features, which are present across all domains, while later layers extract higher-level features, which may differ substantially across domains (such as the people in Fig. \ref{fig:ppl}). Empirically, this is backed up by \cite{wang2019towards}, which shows that activations in later layers differ vastly. 

This approach effectively allows the heads corresponding to smaller domains to learn domain-specific features without suffering from the domain bias induced by the domain imbalances that are favoring larger domains. Note that earlier layers may still be biased towards larger domains. However, as in earlier layers more general-purpose features are extracted \cite{yosinski2014transferable}, this bias is less severe than in later layers.

From preliminary experiments, we found that it is best to split models not based on individual layers, but on groups of layers, which are known as stages or blocks \cite{wang2019towards}. These stages are model-dependent. For example, a Faster-RCNN with a ResNet-101 backbone consists of 5 stages prior to the region-of-interest pooling layer. That means, we share all stages across all domains until a certain stage is reached. From here on, we split the model into so-called experts. For simplicity, these experts are copies of the original model. Therefore, this approach does not need a reorganization of the model architecture and can be applied to many object detectors as will be seen from Section \ref{sec:experiments}.

\begin{figure}
\vspace{2mm}
	
	\centering
	
	\centerline{\includegraphics[width=8.5cm]{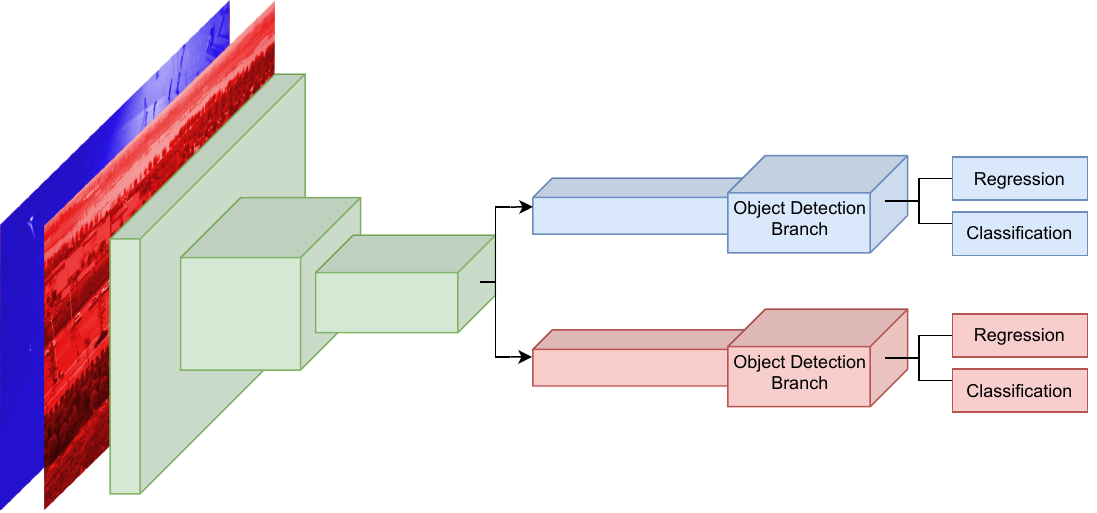}
	}

	%

	\caption{Illustration of a Time@3 model with day and night experts. The time is split into two domains, day (red) versus night (blue), where green outputs represent the shared stages (first, second, third). Every image is passed through the shared green stages. Then it is checked whether it is a day or night image and consequently passed through the red or blue stages, respectively.}
	\label{fig:models}

\end{figure}

For illustrative purposes, see Figure \ref{fig:models}. Here, a Faster-RCNN with ResNet-101 backbone is taken as an example. The first three stages are shared across all domains. Based on the domain label - in this case day or night - the corresponding expert branch is chosen. We denote such a model as Time@3 because the available domains are based on the attribute 'time' and the model is shared until the third stage.

A priori, it is not clear, how many stages should be shared. We explore empirically which stages are to be shared in Section \ref{sec:experiments}.

While the number of parameters scales linearly with the number of domains, the \textit{inference speed stays constant} as only a single expert is evaluated at a time. Therefore, the experts effectively increase a model's capacity without hampering the inference speed. Furthermore, the experts' sizes are still small enough such that they all fit even in embedded GPUs' memory, as will be seen in section \ref{sec:experiments}. 

\subsection{Simplified Training Realization}
\label{sec:training_realization}

So far, the proposed approach may seem as if an adaptation to the model architecture was necessary. However, in the following we want to demonstrate that the expert approach can be implemented in every architecture. Furthermore, it introduces only very little training overhead.

Given an object detector and training pipeline, we train it until an early stopping criterion is met. That means, training it further would increase the validation error. Then, similarly to what is done in transfer learning \cite{zhuang2020comprehensive}, we freeze the shared stages in order to transfer knowledge between domains and such that weights will not be biased towards the over-represented domains \cite{oksuz2020imbalance}. This is particularly beneficial for data sets with great domain imbalances, such as UAVDT and VisDrone. We only train the domain-specific stages further on each respective domain. We split a subset from the training set for that particular domain and use it as the validation set. We train until the validation error increases again. Finally, we take the weights corresponding to the lowest validation loss as our final weights for that expert. Even though the number of trainable parameters shrinks, we want to emphasize that having a validation set is especially critical in this case to avoid overfitting on the small domains. 


Post-training the domain-specific layers on their corresponding domains introduces little overhead to the overall pipeline. This is because only a small number of layers needs to be trained which decreases the time for the backward pass because only parts of the weights need to be back-propagated and the freed GPU memory space can be used to increase the batch size. Furthermore, training for different domains can be done in parallel. We report actual training times for various experiments in Section \ref{sec:experiments}.

\subsection{Introducing a Multi-Modal Data Set}
\label{sec:multi-modal-dataset}

Lastly, we would like to note that there are no publicly available data sets for object detection from UAVs that include precise domain labels regarding altitude and viewing angle. E.g. \cite{bozcan2020air} includes limited altitude values between 0-30m. We argue that this is a major impediment in the development of domain-aware models since these two factors majorly contribute to appearance changes. 

\begin{figure}

	\centering
	
    \centerline{\includegraphics[trim={0 0 2cm 0},clip,width=8.5cm]{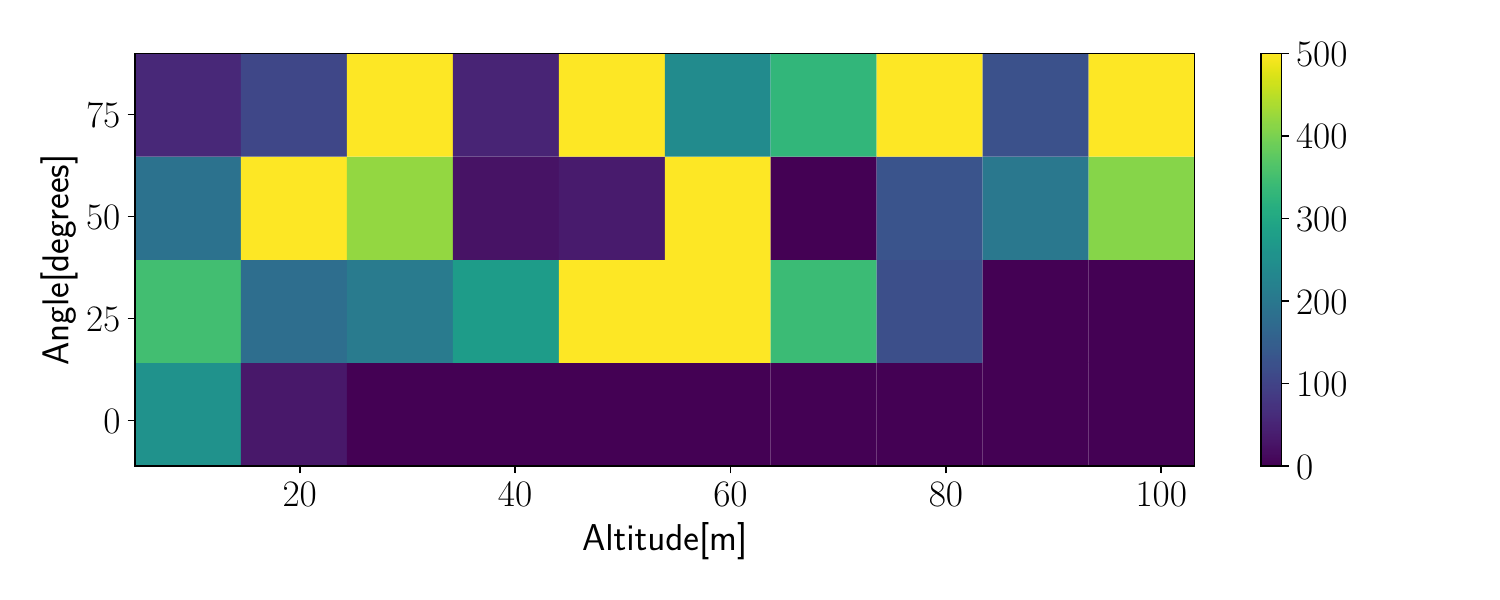}
	}

	%

	\caption{Distribution of objects in PeopleOnGrass (POG) across different levels of altitude and camera pitch angles. For visualization purposes only a 4x10 grid is shown.}
	\label{fig:pog_hist}

\end{figure}

For that reason, we record the experimental data set PeopleOnGrass (POG) containing 2,9k images (3840x2160 pixels resolution), showing people from various angles and altitudes varying from  0$^{\circ}$ (horizontally facing) to 90$^{\circ}$ (top-down) and 4m to 103m, respectively, each labeled with the precise altitude and angle it was captured at. See Figure \ref{fig:pog_hist} for a distribution of objects. Further metadata, such as GPS location, UAV speed and rotation, timestamps and others are also included. 
We use a DJI Matrice 210 equipped with a Zenmuse XT2. The meta data is obtained through DJI's onboard software developing kit. Accompanied with every frame there is a meta stamp, that is logged at 10 hertz. To align the video data (30 fps) and the time stamps, a nearest neighbor method was performed. The following data is logged and provided for every image/frame read from the onboard clock, barometer, IMU and GPS sensor, respectively:

\begin{itemize}
	\item current date and time of capture
	\item latitude, longitude and altitude of the UAV
	\item camera pitch, roll and yaw angle (viewing angle)
	\item speed along the $x$-, $y$ and $z$-axes
\end{itemize}

We need to emphasize that the the meta values lie within the error thresholds introduced by the different sensors but an analysis is beyond the scope of this paper (see \cite{webinar2020gps} for an overview).

See Figure \ref{fig:ppl} for example images. We manually and carefully annotate 13,713 people. We note that this is a simple real-world data set, suffering from fewer confounders than large data sets which is ideal for testing out the efficacy of multi-modal methods.

This data set will be released with the paper and hopefully will benefit the development of multi-modal models.

\begin{table}
	\caption{Several domain expert results for various freezing strategies on VisDrone. Altitude@x means that all stages until the $x$th. stage are shared.}
	\begin{center}		
	\setlength\tabcolsep{2.5pt}
			\begin{tabular}{c|ccccccc}
				
				& L & M & H & mAP$_{50}$ & mAP & mAP$_{50}^\text{avg}$ & T\\
				\hline
				DE-FPN \cite{zhu2018visdrone} & 49.1 & 49.7 & 36.0 & 48.6& 26.1 &44.9 & --\\
				
				Altitude@0&  49.4 & 49.6 & 35.5 & 48.3 & 25.9& 44.8 & 12 \\
				
				Altitude@1& 49.5 & 49.7 & 35.7 & 48.5 & 25.9& 45.0 & 11 \\
				
				Altitude@2& 49.5 & 49.9 & 36.1 & 48.7 & 26.1& 45.2 & 11 \\
				
				Altitude@3& 50.2 & \bf50.2 & 36.8 & 49.2 & 26.6& 45.7 & 10 \\
				
				Altitude@4& \bf 50.7 & \bf50.2 & \bf37.5 & \bf49.9 & \bf27.4& \bf46.1 & 8 \\	
				
				Altitude@5&  50.5 & 50.0 & \bf37.5 & 49.7 & 27.0& 46.0 & 7 \\
				\hline
				\hline
				& B & A & &  &  & \\
				\hline
				DE-FPN \cite{zhu2018visdrone} & 38.0 & 49.0 &  & 48.6& 26.1 & 43.5 & --\\
				
				Angle@4& \bf 39.6 & \bf49.8 &  & \bf49.4 & \bf27.0& \bf44.7 & 6\\	

				\hline
				\hline
				& D & N & &  &  & \\
				\hline
				DE-FPN \cite{zhu2018visdrone} & 48.5 & 52.0 &  & 48.6& 26.1 & 50.2 & -- \\
				
				Time@4& \bf 49.0 & \bf52.6 &  & \bf49.0 & \bf26.6& \bf50.8 & 7\\

		\end{tabular}
	\end{center}

	\label{table:visdrone-freeze}
\end{table}

\section{Experimental Results and Ablations}
\label{sec:experiments}

In the first two sets of experiments, we show how leveraging domain labels on UAVDT and VisDrone improves multiple model architectures' performances. Furthermore, we investigate the effect of different splitting strategies and ablations. Lastly, we show that finer domain splitting is possible in the case of the data set POG.

\begin{table*}
    \vspace{2mm}
	\caption{Results on specific domains for multi-dimension experts on VisDrone. For example, the Altitude-angle-time@4-expert achieves 54.4 mAP$_{50}$ on the domain L+A+N (low altitude, acute viewing angle, and at night).  }
	\begin{center}
		
			\begin{tabular}{c|c|ccc|cccc}
								& $\downarrow$ + $\rightarrow$ & L & M & H & mAP$_{50}$ & mAP & mAP$_{50}^{\text{avg}}$ & T\\ \hline 
					DE-FPN \cite{zhu2018visdrone}      & \begin{tabular}
						{c}   B \\ A
					\end{tabular} &  \begin{tabular}
					{c}   84.6 \\ 49.1
				\end{tabular} & \begin{tabular}
				{c}   42.5 \\ 50.0
			\end{tabular} & \begin{tabular}
			{c}   35.6 \\  41.2
		\end{tabular} & 48.6 & 26.1 & 50.5 & --\\ \hdashline

				\begin{tabular}{c}
					Altitude-angle@4
				\end{tabular}       & \begin{tabular}
					{c} B \\ A 
				\end{tabular}  & \begin{tabular}
					{c} \bf 87.4  \\ \bf 49.7	\end{tabular} & 
				
				\begin{tabular}
					{c} \bf 44.8  \\ \bf 50.1	\end{tabular} & 
				
				\begin{tabular}
					{c} \bf 39.6  \\ \bf 42.2	\end{tabular} & \bf 49.0 & \bf 26.3 &\bf  52.3 & 10\\ \hline \hline

				   DE-FPN \cite{zhu2018visdrone}       & \begin{tabular}
					{c} B+D \\ A+D \\ A+N
				\end{tabular}  & \begin{tabular}
					{c} 84.6 \\ 49.0 \\ 52.8		\end{tabular} &
				\begin{tabular}
					{c} 42.5 \\ 50.2 \\ 51.6		\end{tabular}
				&
				\begin{tabular}
					{c} 35.6 \\ 41.2 \\ --	\end{tabular} & 48.6 & 26.1 & 50.9& --
				 \\\hdashline
				\begin{tabular}{c}
					Altitude-angle-time@4
				\end{tabular}       & \begin{tabular}
				 	{c} B+D \\ A+D \\ A+N 
				 \end{tabular}  & \begin{tabular}
				 	{c} \bf 87.5  \\ \bf 50.1 \\ \bf 54.4	\end{tabular} & 
				 
				 \begin{tabular}
				 	{c} \bf 44.8  \\ \bf 50.6 \\ \bf 56.5	\end{tabular} & 
				 
				 \begin{tabular}
				 	{c} \bf 39.6  \\ \bf 42.2 \\ --	\end{tabular} & \bf 49.6 & \bf 26.8 &\bf  52.9 & 11
		\end{tabular}
	\end{center}

	\label{table:visdrone-multi-altitude-angle}
\end{table*}

\subsection{VisDrone}
\label{sec:visdrone}

We evaluate our models using the official evaluation protocols, i.e. AP$_{70}$ for UAVDT and mAP and mAP$_{50}$ for VisDrone, respectively. Furthermore, we report results on individual domains and the domain-averaged metric from Section \ref{sec:metric}, i.e. AP$_{70}^\text{avg}$ and mAP$_{50}^\text{avg}$ over all respective domains to measure the universal cross-domain performance. The subscript $50$ and $70$ denote the intersection-over-union (IoU) a prediction needs to have with a ground truth bounding box in order to be counted as a true positive. Note that we leave out the 'm' in 'mAP' for UAVDT since it contains only one class. 

Furthermore, we report the additional training times $T$ in percent (rounded to integers) to train a model longer than its baseline, i.e. $T=10$ would mean that it takes additional 10\% to train a model further than its baseline.

The object detection track from VisDrone consists of around 10k images with 10 categories. All frames are annotated with domain labels regarding altitude (low (L), medium (M), high (H)), viewing angle (front, side, bird (B)) and light condition (day (D), night (N)) \cite{wu2019delving}. Note that we fuse the two domains "front" and "side" into a single domain "acute angle (A)", as, at test time, we can only distinguish between bird view and not bird view based on the camera angle.
We reimplement the best performing single-model (no ensemble) from the workshop report, DE-FPN \cite{zhu2018visdrone}, i.e. a Faster R-CNN with a ResNeXt-101 64-4d \cite{xie2017aggregated} backbone (removing P6), which was trained using color jitter and random image cropping achieving 48.7\% mAP$_{50}$ on the test set. To compare with \cite{wu2019delving}, we evaluate our models on the unseen validation set, on which our implementation yields 48.6\% mAP$_{50}$. 


 From Table \ref{table:visdrone-freeze}, we can make four observations: First, the altitude-experts improve over the baseline DE-FPN on the whole validation set and on all domains individually if more than up until the second stage is shared. The performance drop of Altitude@0 and Altitude@1 is likely caused by overfitting on the small domain H, on which the performance drop is -0.5 mAP$_{50}$. Note that Altitude@0 essentially has a separate model for each domain.  Second, there seems to be an upward trend in performance, peaking at the fourth stage and dropping at the fifth stage. Third, improvements are seen for all experts: +1.3, +0.8 and +0.4 mAP$_{50}$ for the \hbox{Altitude-,} Angle- and Time-experts, respectively. Furthermore, the performance improvements are also seen in the domain-sensitive metric mAP$_{50}^{\text{avg}}$, yielding +1.2, +1.2 and +0.6 points for the respective experts. Lastly, the additional training time is low, with 8\%, 6\% and 7\% for the most accurate domain experts. As it yielded the best results, we always freeze until the $4$th stage for VisDrone from here on.

Table \ref{table:visdrone-multi-altitude-angle} shows that sharing along two and three domain dimensions is advantageous. The Altitude-angle@4-experts and the Altitude-angle-time@4-experts improve DE-FPN on all domains individually and overall. In particular, we obtain a +1.8 and +2 mAP$_{50}^\text{avg}$ increase, respectively. The standard metrics mAP and mAP$_{50}$ show an improvement as well, albeit a lower one which is attributed to the failure of these metrics to capture domain imbalances in the validation set (see Figure \ref{table:visdrone-count}). 

This contrast is, furthermore, seen in underrepresented domains being improved the most, suggesting the diminished domain bias. For example, the Altitude-angle-time@4-experts improve the performance on the domains M+A+N and H+B+D, which only contain roughly 4\% and 8\% of all objects (see Figure \ref{table:visdrone-count}), from 51.6 mAP$_{50}$ to 56.5 mAP$_{50}$ and 35.6 mAP$_{50}$ to 39.6 mAP$_{50}$, respectively.

\begin{table}
	\caption{Altitude-time@4 and Angle-time@4 experts on the VisDrone validation set.}
	\begin{center}		
	
			\begin{tabular}{c|cccc}
				
			 & mAP$_{50}$ & mAP & mAP$_{50}^\text{avg}$ & T\\
			
				\hline

				DE-FPN \cite{zhu2018visdrone} & 48.6 & 26.1 & 49.7 & --\\
				
				Altitude-time@4& \bf 49.1 & \bf26.3 &  \bf51.5 & 11\\	
				
				\hline
				\hline

				DE-FPN \cite{zhu2018visdrone} & 48.6 & 26.1 & 50.1 & -- \\
				
				Angle-time@4& \bf 49.2 & \bf26.4 & \bf 51.9 & 13\\

			\end{tabular}
	\end{center}

	\label{table:multi-exp-visdrone}
\end{table}

Similar observations can be made from Table \ref{table:multi-exp-visdrone}, where the Altitude-time@4- and Angle-time@4-experts both improve by +1.8 mAP$_{50}^{\text{avg}}$.

\begin{table}
\vspace{1.5mm}
	\caption{EfficientDet-D0 Angle experts on VisDrone validation set}
	\begin{center}

			\begin{tabular}{c|cccc}
				
				& B & A & mAP$_{50}$  & mAP$_{50}^\text{avg}$\\
				\hline
				EfficientDet-D0 & 21.5 & 24.9 & 26.3 & 23.2  \\

				Angle@backbone &  \bf 22.1 &  \bf 26.2 & \bf 27.6 & \bf 24.2 				
				
		\end{tabular}
	\end{center}

	\label{table:efficient-visdrone}
\end{table}

To further test our approach in real-time scenarios, we choose the current best model family on the COCO test-dev according to \cite{paperswithcode}, i.e. EfficientDet \cite{tan2020efficientdet}, and take the smallest model D0 as our baseline model. We employ it on the NVIDIA Jetson AGX Xavier suitable for on-board processing \cite{ditty2018nvidia}. For that, we convert the trained model to half-precision using JetPack and TensorRT \cite{vanholder2016efficient} and set the performance mode to MAX-N. The inference speed is reported in frames per second (fps) averaged over the validation set. Similar to \cite{ringwald2019uav}, the fps values do not include the non-maximum suppression stage as TensorRT does not supported it yet. Keeping the image ratio, the employed longer image side is 1408 pixels for training and testing. 

We freeze the whole backbone and only leave the box-prediction net \cite{tan2020efficientdet} domain-specific. As shown in Table \ref{table:efficient-visdrone}, sharing the backbone yields an improvement of 1.3 point mAP$_{50}$ and 1 point mAP$_{50}^\text{avg}$ for the angle experts. Both models run at 21.8fps, suitable for live on-board processing. With all pre- and post-processing steps, we obtain a wall-clock time of 18.1fps.

\subsection{UAVDT}
\label{sec:uavdt}

\begin{table}	
	\begin{center}
	\caption{Domain experts on the UAVDT testing set.}
	\setlength\tabcolsep{3pt}
			\begin{tabular}{c|cccccc}
				
			& L & M & H & AP$_{70}$ & AP$_{70}^\text{avg}$ & T\\
				\hline
				ResNet-101-FPN \cite{wu2019delving} & 61.9 & 58.1 & \bf 24.1 & \bf 49.4& 48.0 & -- \\
				Altitude@2& \bf 62.5 & \bf60.5 & \bf24.1 & \bf49.4 & \bf49.0 & 10\\
				
				\hline
				\hline
				& B & A & &  &   \\
				\hline
				ResNet-101-FPN \cite{wu2019delving} & 28.9 & 59.1 &  & 49.4& 44.0 & -- \\
				Angle@2 & \bf33.6 & \bf60.4 &  & \bf50.4& \bf 47.0  & 9\\
				
				\hline
				\hline
				& D & N & &  &   \\
				\hline
				ResNet-101-FPN \cite{wu2019delving} & 51.4 & 50.6 &  & 49.4 & 51  & --\\
				
				Time@2& \bf 53.4 & \bf 54.1 &  & \bf50.1 & \bf53.8 & 10\\
				
		\end{tabular}
	\end{center}

	\label{table:uavdt-single}
\end{table}

UAVDT contains around 41k annotated frames with cars, busses and trucks. Similar to \cite{wu2019delving}, we fuse all classes into a single vehicle class. All frames are domain-annotated like VisDrone.
To compare our experts, we trained a Faster R-CNN with ResNet-101-FPN similar to \cite{wu2019delving}, which report 49.1 AP$_{70}$ on the testing set. We obtain 49.4 AP$_{70}$ on the testing set and we compare with that value. 

As Table VI shows, the Angle@2- and Time@2-experts improve performance over the baseline on both metrics. In particular, the Angle@2-expert improves the baseline by +3 points AP$_{70}^\text{avg}$. Furthermore note, that there is not a accuracy increase in domain H, since there are almost no training images available ($\approx 1\%$).

\begin{table}

	\begin{center}
	\caption{Results for ResNet-101 backbone on UAVDT}

			\begin{tabular}{c|cccc}
				
			& B & A  & AP$_{70}$ & AP$_{70}^\text{avg}$\\
				
				\hline				
				ResNet-101 \cite{wu2019delving} & 27.1 & 54.4 &   45.6& 40.1  \\
				NDFT \cite{wu2019delving} & 28.8 & 56.0 &   47.9&  43.4  \\
				Angle@2 & \bf31.6 & \bf58.6 &   \bf48.6& \bf 45.1

		\end{tabular}
	\end{center}

	\label{table:uavdt-ndft}
\end{table}

We also demonstrate that the performance gain using expert models does not vanish as we switch to another backbone, e.g. ResNet-101. As shown in Table VII, the angle experts yield an increase in +3 AP$_{70}$ and +5 AP$_{70}^\text{avg}$ and even outperform NDFT \cite{wu2019delving}, an approach using adversarial losses on domain labels.

\begin{table}
\vspace{1.5mm}
    	
	\label{table:efficient-uavdt}
	\begin{center}
    \caption{Altitude experts results on UAVDT test set}

			\begin{tabular}{c|ccc}
				
			    & AP$_{70}$ & FPS & AP$_{70}^\text{avg}$ \\
				\hline
				EfficientDet-D0 & 17.1 & \bf 21.8 & 16.7 \\
				{UAV-Net \cite{ringwald2019uav}} & 26.2 & 18.3 & -- \\ 
				Altitude@backbone & \bf 38.1 & \bf 21.8 & \bf 37.0 \\
				
		\end{tabular}
	\end{center}

\end{table}

Finally, we also test a real-time detector on UAVDT. Similar as for VisDrone, Table VIII shows how the altitude experts with shared backbone can regain precision that has been sacrificed to the high speed of the D0 model. The large improvement of +21.0 AP$_{70}$ is likely caused by the domain bias induced by the heavy altitude imbalance of UAVDT (see Table \ref{table:visdrone-count}), which the experts are successful to mitigate. 

In particular, we set a new state-of-the-art performance for real-time detectors on embedded hardware by improving upon \cite{ringwald2019uav} by +11.9 AP$_{70}$ while being 3.5fps faster. Note that they tested on different embedded hardware.

\begin{table}

	\begin{center}
	\caption{(Finer) Altitude experts results on POG test set }
			\begin{tabular}{c|ccc}
				
				&  AP$_{50}$ & AP & AP$_{50}^{\text{avg}}$\\
				\hline
				EfficientDet-D0 & 82.0 & 36.4  &82.9\\				
				3xAltitude@backbone&  86.2 & 40.3 & 86.0  \\
				
				6xAltitude@backbone& \bf 87.9 & \bf40.8 & \bf88.1  \\

		\end{tabular}
	\end{center}

	\label{table:pog}
\end{table} 

\subsection{POG: Baseline and Expert Results}
\label{sec:ppl}

Finally, we test the expert approach on our own captured data set POG. For future reference, we establish an EfficientDet-D0 baseline which can run in real-time on embedded hardware such as the Xavier board. Finally, we employ altitude experts with shared backbone to showcase the effectiveness of a multi-domain learning approach on finer domains. 

We split the altitude range (4m -- 103m) into three and six \textit{equidistant} domains, respectively. That is, our domains are 

\begin{enumerate}
\item $d_1=(4,37)$, $d_2=[37,70)$, $d_3=[70,103)$
\item $d_1=(4,20.5)$, $d_2=[20.5,37)$, $d_3=[37,53.5)$,\\ $d_4=[53.5,70)$, $d_5=[70,86.5)$, $d_6=[86.5,103)$,
\end{enumerate}

respectively. We denote the corresponding experts as 3xAltitude (1.) and 6xAltitude (2.), respectively. As before, we freeze the backbone and report results for the fast EfficientDet-D0. 
Table IX shows that the baseline achieves 82.0 AP$_{50}$, which the experts improve by +4.2 and +5.9 AP$_{50}$, respectively, showing that experts further benefit from finer domain splits (6xAltitude +1.7 AP$_{50}$ compared to 3xAltitude). See the video for a visual demonstration of an expert model.

\section{Conclusion and Limitations}
\label{sec:conclusion}

We successfully apply a multi-domain learning method to object detection from UAVs. We propose and analyze expert models, leveraging domain data at test time. Although these expert models are conceptually simple, they achieve domain awareness and consistently improve several, heavily optimized state-of-the-art models on multiple data sets and metrics. In particular, our EfficientDet-D0 altitude expert yields 38.1\% AP$_{70}$ on UAVDT, making it the new state-of-the-art real-time detector on embedded hardware. 

However, we believe that domain labels in UAV object detection can be exploited even more. In particular, the assumption that domains are regarded as equally discrete may be overly strict. In future work, it would be worth investigating how domains interact with each other on a deeper level. In this matter, incorporating softer boundaries between domains could be a promising direction. Furthermore, different sampling strategies, such as oversampling small domains, could be investigated.

\bibliographystyle{IEEEtran.bst} 
\bibliography{IEEEabrv,root}

\end{document}